\documentclass{article}

\PassOptionsToPackage{numbers, compress}{natbib}
\usepackage[final]{nips_2016}

\usepackage[utf8]{inputenc} 
\usepackage[T1]{fontenc}    
\usepackage{hyperref}       
\usepackage{url}            
\usepackage{booktabs}       
\usepackage{amsfonts}       
\usepackage{nicefrac}       
\usepackage{microtype}      
\usepackage{graphicx}
\usepackage{afterpage}
\usepackage{amsmath}
\usepackage{mathtools}
\usepackage{setspace}

\title{Texture Synthesis Using Shallow Convolutional Networks with Random Filters}

\author{
  Ivan Ustyuzhaninov\textsuperscript{*,1,2,3}, Wieland Brendel\textsuperscript{*,1,2}, Leon Gatys\textsuperscript{1,2,3}, Matthias Bethge\textsuperscript{1,2,3,4}  \\[0.5cm]
  \textsuperscript{*}contributed equally\\
\textsuperscript{1}Centre for Integrative Neuroscience, University of Tübingen, Germany\\  \textsuperscript{2}Bernstein Center for Computational Neuroscience, Tübingen, Germany\\
  \textsuperscript{3}Graduate School of Neural Information Processing, University of Tübingen, Germany \\
  \textsuperscript{4}Max Planck Institute for Biological Cybernetics, Tübingen, Germany \\[0.25cm]
  \texttt{first.last@bethgelab.org} \\
}

\begin{document}

\maketitle

\begin{abstract}
Here we demonstrate that the feature space of random shallow convolutional neural networks (CNNs) can serve as a surprisingly good model of natural textures. Patches from the same texture are consistently classified as being more similar then patches from different textures. Samples synthesized from the model capture spatial correlations on scales much larger then the receptive field size, and sometimes even rival or surpass the perceptual quality of state of the art texture models (but show less variability). The current state of the art in parametric texture synthesis relies on the multi-layer feature space of deep CNNs that were trained on natural images \citep{gatys:2015a}. Our finding suggests that such optimized multi-layer feature spaces are not imperative for texture modeling. Instead, much simpler shallow and convolutional networks can serve as the basis for novel texture synthesis algorithms.
\end{abstract}

\section{Introduction}

The aim of visual texture synthesis is to define a generative process that, from a given example texture, can generate arbitrarily many new samples of the same texture. Among the class of such algorithms, parametric texture models aim to uniquely describe each texture by a set of statistical measurements that are taken over the spatial extent of the image. Each image with the same spatial summary statistics should be perceived as the same texture. Consequently, synthesizing a texture corresponds to finding a new image that reproduces the summary statistics of the reference texture. Starting from Nth-order joint histograms of the pixels by Julesz \cite{julesz1962}, many different statistical measures have been proposed \cite[see e.g.][]{Heeger1995,Portilla:2000}. The quality of the synthesized textures is usually determined by human inspection; the synthesis is successful if a human observer cannot tell the reference texture from the synthesized ones.

The current state of the art in parametric texture modeling \cite{gatys:2015a} employs the hierarchical image representation in a deep 19-layer convolutional network (in the following referred to as VGG network) \cite{VGG:2014} that was trained on object recognition in natural images. In this model textures are described by the raw correlations between feature activations in response to the texture image from a collection of network layers (see section \ref{sec:textureeval} for details). Two aspects of the model seemed critical for texture synthesis: the hierarchical multi-layer representation of the textures, and the supervised training of the feature spaces. Here we show that neither aspect is imperative for texture modeling and that in fact a single convolutional layer with random features can often synthesize textures that rival, and sometimes even surpass, the perceptual quality of Gatys et al. \cite{gatys:2015a}. This is in contrast to Gatys et al. \cite{gatys:2015a} who reported that networks with random weights fail to generate perceptually interesting images. We suggest that this discrepancy originates from a more elaborate tuning of the optimization procedure (see section \ref{sec:texturesynth}).

\section{Convolutional Neural Network}
\label{sec:cnn}

All our models employ single-layer CNNs with standard rectified linear units (ReLUs) and convolutions with stride one, no bias and padding $(f-1)/2$ where $f$ is the filter-size. This choice ensures that the spatial dimension of the output feature maps is the same as the input. All networks except the last one employ filters of size $11\times 11\times 3$ (filter width $\times$ filter height $\times$ no. of input channels), but the number of feature maps as well as the selection of the filters differ:

\begin{itemize}
	\item \textbf{Fourier-363: } Each color channel (R, G, B) is filtered separately by each element ${\bf B}_i\in\mathbb{R}^{11\times 11}$ of the 2D Fourier basis ($11\times 11 = 121$ feature maps/channel), yielding $3\cdot 121 = 363$ feature maps in total. More concretely, each filter can be described as the tensor product ${\bf B}_i\otimes {\bf e}_k$ where the elements of the unit-norm ${\bf e}_k\in\mathbb{R}^{3}$ are all zero except one.
	\item \textbf{Fourier-3267: } All color channels (R, G, B) are filtered simultaneously by each element ${\bf B}_i$ of the 2D Fourier basis but with different weighting terms $w_R, w_G, w_B \in [1, 0, -1]$, yielding $3\cdot 3\cdot 3\cdot 121 = 3267$ feature maps in total.  More concretely, each filter can be described by the tensor product ${\bf B}_i \otimes [w_R, w_G, w_B]$.
	\item \textbf{Kmeans-363: } We randomly sample and whiten 1e7 patches of size $11\times 11$ from the Imagenet dataset \cite{ILSVRC15}, partition the patches into 363 clusters using k-means \cite{rubinstein:2009}, and use the cluster means as convolutional filters.
	\item \textbf{Kmeans-3267: } Same as Kmeans-363 but with 3267 clusters.
	\item \textbf{Kmeans-NonWhite-363/3267: } Same as Kmeans-363/3267 but without whitening of the patches.
	\item \textbf{Kmeans-Sample-363/3267: } Same as Kmeans-363/3267, but patches are only sampled from the target texture.
	\item \textbf{PCA-363: }  We randomly sample 1e7 patches of size $11\times 11$ from the Imagenet dataset \cite{ILSVRC15}, vectorize each patch, perform PCA and use the set of principal axes as convolutional filters.
	\item \textbf{Random-363: } Filters are drawn from a uniform distribution according to \citep{glorot:2010}, 363 feature maps in total.
	\item \textbf{Random-3267: } Same as Random-363 but with 3267 feature maps.
	\item \textbf{Random-Multiscale} Eight different filter sizes $f\times f\times 3$ with $f = 3, 5, 7, 11, 15, 23, 37, 55$ and 128 feature maps each (1024 feature maps in total). Filters are drawn from a uniform distribution according to \citep{glorot:2010}.
\end{itemize}

The networks were implemented in Lasagne \citep{lasagne, theano:2016}. We remove the DC component of the inputs by subtracting the mean intensity in each color channel (estimated over the Imagenet dataset \cite{ILSVRC15}).

\section{Texture Model}
\label{sec:texturemodel}
\begin{figure}[t]
  \centering
      \makebox[\textwidth][c]{\includegraphics[width=1\textwidth]{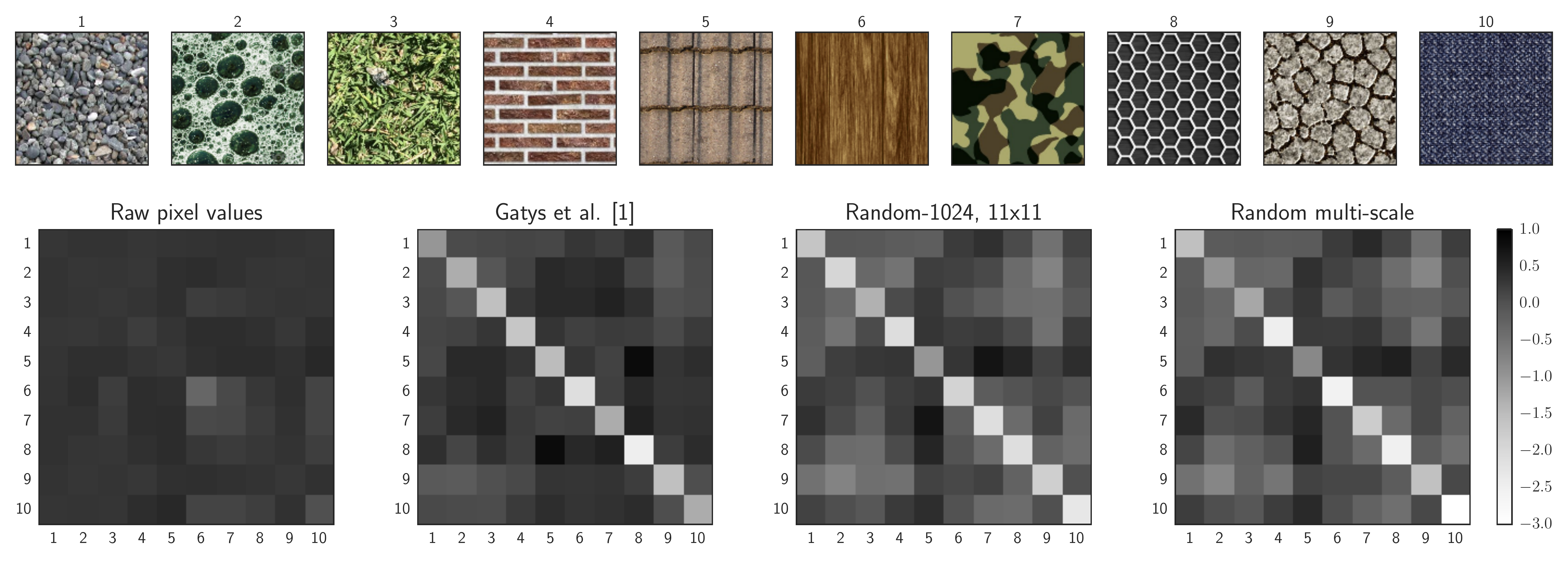}}
  \caption{Similarity measures between textures for three different texture models as well as raw pixel values. Ten random patches were extracted from each of ten different textures (examples of patches are shown above the matrices), and normalized Euclidean distances \eqref{eq:gramdist} (\eqref{eq:pixeldist} for raw pixel values) between patches from all pairs of textures were computed in representation spaces of three different models (VGG, single-layer net with random 11x11 filters and single-layer net with random multi-scale filters) as well as the raw pixel values (left). The matrix element $(i, j)$ corresponds to the median distance between patches from textures $i$ and $j$. The values in this figure are shown on a log-scale.}
  \label{fig:confusion}
\end{figure}
The texture model closely follows \citep{gatys:2015a}. In essence, to characterise a given vectorised texture ${\bf x} \in\mathbb{R}^M$, we first pass $\bf x$ through the convolutional layer and compute the output activations. The output can be understood as a non-linear filter bank, and thus its activations form a set of filtered images (so-called feature maps). For $N$ distinct feature maps, the rectified output activations can be described by a matrix ${\bf F}\in\mathbb{R}^{N\times M}$. 
To capture the stationary structure of the textures, we compute the covariances (or, more precisely, the Gramian matrix) ${\bf G}\in\mathbb{R}^{N\times N}$ between the feature activations $\bf F$ by averaging the outer product of the point-wise feature vectors,
\begin{equation}
	\label{eq:gram}
	G_{ij} = \frac{1}{M} \sum_{m = 1}^{M} {F}_{im} {F}_{jm}.
\end{equation}
We will denote ${\bf G}({\bf x})$ as the Gram matrix of the feature activations for the input ${\bf x}$. To determine the relative distance between two textures ${\bf x}$ and ${\bf y}$ we compute the euclidean distance of the normalized Gram matrices,
\begin{equation}
	\label{eq:gramdist}
	d({\bf x}, {\bf y}) = \frac{1}{\sqrt{\sum\limits_{m,n} G_{mn}({\bf x})^2}\sqrt{\sum\limits_{m,n} G_{mn}({\bf y})^2}}\sum_{i,j=1}^{N}\left(G_{ij}({\bf x}) - G_{ij}({\bf y})\right)^2.
\end{equation}
To compare with the distance in the raw pixel values, we compute
\begin{equation}
	\label{eq:pixeldist}
	d_p({\bf x}, {\bf y}) = \frac{1}{\sqrt{\sum\limits_{m} x_m^2}\sqrt{\sum\limits_{m} y_m^2}}\sum_{i=1}^{N}\left(x_i - y_i)\right)^2.
\end{equation}

\section{Texture Synthesis}
\label{sec:texturesynth}

To generate a new texture we start from a uniform noise image (in the range [0, 1]) and iteratively optimize it to match the Gram matrix of the reference texture. More precisely, let ${\bf G}({\bf x})$ be the Gram matrix of the reference texture. The goal is to find a synthesised image ${\bf \tilde x}$ such that the squared distance between ${\bf G}({\bf x})$ and the Gram matrix ${\bf G}({\bf \tilde x})$ of the synthesized image is minimized, i.e.
\begin{align}
	{\bf \tilde x} &= \underset{{\bf y}\in\mathbb{R}^M}{\operatorname{arg min}}\,\,E({\bf y}), \label{eq:minx} \\
    E({\bf y}) &= \frac{1}{\sum_{i,j=1}^{N}G_{ij}({\bf x})^2}\sum_{i,j=1}^{N}\Big(G_{ij}({\bf x}) - G_{ij}({\bf y})\Big)^2. \label{eq:synthdist}
\end{align}
The gradient $\partial E({\bf y}) / \partial {\bf y}$ of the reconstruction error with respect to the image can readily be computed using standard backpropagation, which we then use in conjunction with the L-BFGS-B algorithm \cite{scipy:2001} to solve \eqref{eq:minx}. We leave all parameters of the optimization algorithm at their default value except for the maximum number of iterations (2000), and add a box constraints with range [0, 1]. In addition, we scale the loss and the gradients by a factor of $10^7$ in order to avoid early stopping of the optimization algorithm (which might have caused the negative results for random networks in \cite{gatys:2015a}).
\begin{figure}[t]
  \centering
      \makebox[\textwidth][c]{\includegraphics[width=1\textwidth]{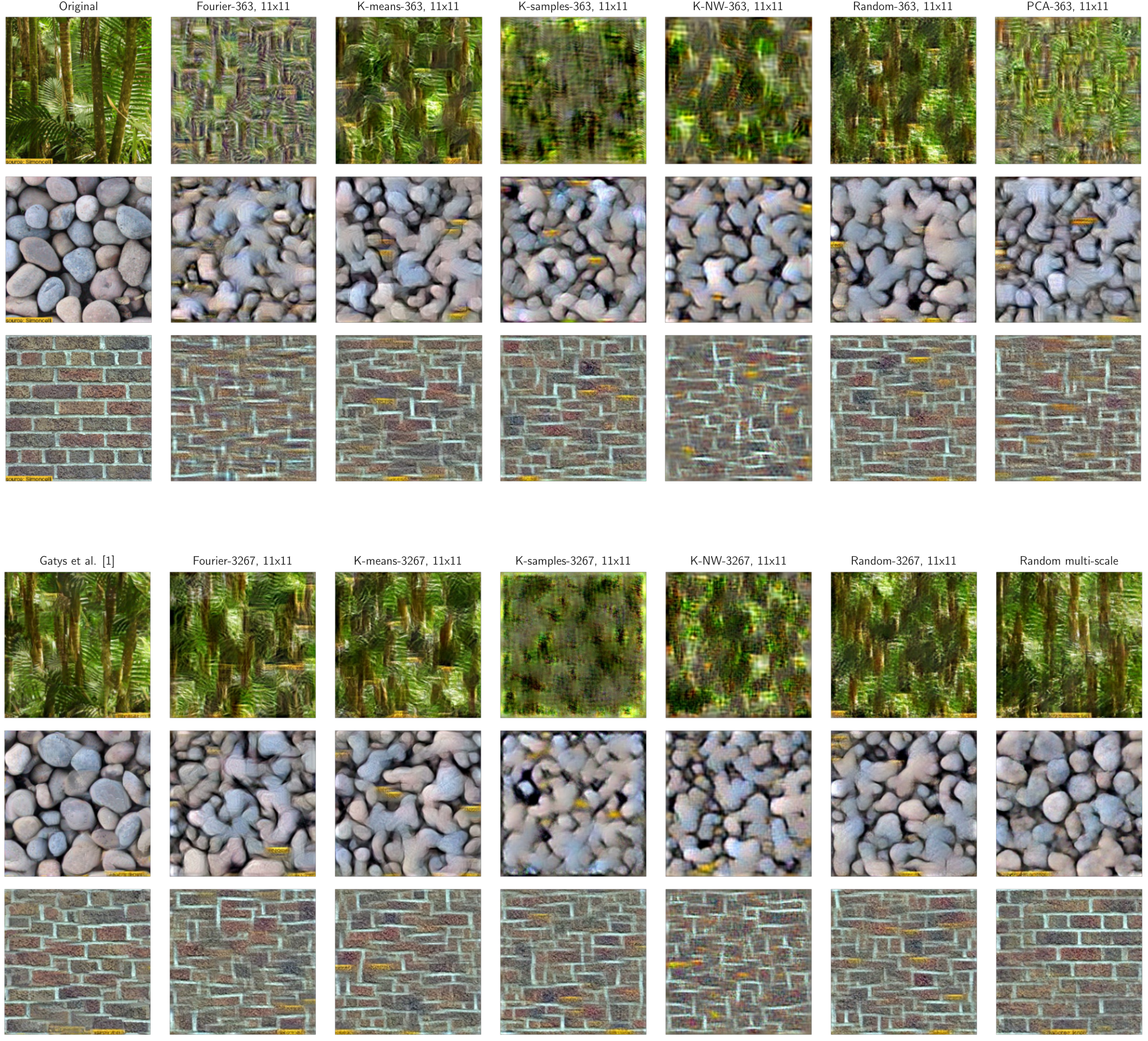}}
  \caption{(Top) Samples synthesized from several single-layer models with 363 feature maps (see sec. \ref{sec:cnn}) for three different textures (rows). Reference textures are shown in the first column. (Bottom) Samples synthesized from several single-layer models with 3267 feature maps (see sec. \ref{sec:cnn}) for three different textures (rows). Additionally, the first column shows samples from the VGG model \cite{gatys:2015a}, and the last column from the multi-scale model (with 1024 feature maps).}
  \label{fig:filter-samples}
\end{figure}
\section{Texture Evaluation}
\label{sec:textureeval}

Evaluating the quality of the synthesized textures is traditionally performed by human inspection. Optimal texture synthesis should generate samples that humans perceive as being the same texture as the reference. The high quality of the synthesized textures by \citep{gatys:2015a} suggests that the summary statistics from multiple layers of VGG can approximate the perceptual metric of humans. Even though the VGG texture representation is clearly not perfect (see Fig. \ref{fig:confusion}), this allows us to utilize these statistics as a more objective quantification of texture quality.

For all details of the VGG-based texture model see \citep{gatys:2015a}. Here we use the standard 19-layer VGG network \cite{VGG:2014} with pretrained weights and average- instead of max-pooling\footnote{https://github.com/Lasagne/Recipes/blob/master/modelzoo/vgg19.py as accessed at 12.05.2016.}. We compute a Gram matrix on the output of each convolutional layer that follows a pooling layer. Let ${\bf G}^\ell(.)$ be the Gram matrix on the activations of the $\ell$-th layer and
\begin{equation}
    E^\ell({\bf y}) = \frac{1}{\sum_{i,j=1}^{N}G^\ell_{ij}({\bf x})^2}\sum_{i,j=1}^{N}\Big(G^\ell_{ij}({\bf x}) - G^\ell_{ij}({\bf y})\Big)^2.
\end{equation}
the corresponding relative reconstruction cost. The total reconstruction cost is then defined as the average distance between the reference Gram matrices and the synthesized ones, i.e.
\begin{equation}
	\label{eq:VGGloss}
	E({\bf y}) = \frac{1}{5}\sum_{\ell = 1}^{5}E^{\ell}({\bf y}).
\end{equation}
This cost is reported on top of each synthesised texture in Figures \ref{fig:samples}. To visually evaluate samples from our single- and multi-scale model against the VGG-based model \cite{gatys:2015a}, we additionally synthesize textures from VGG by minimizing \eqref{eq:VGGloss} using L-BFGS-B as in section \ref{sec:texturesynth}.

\section{Results}
\label{sec:results}
\begin{figure}[t]
  \centering
      \makebox[\textwidth][c]{\includegraphics[width=0.7\textwidth]{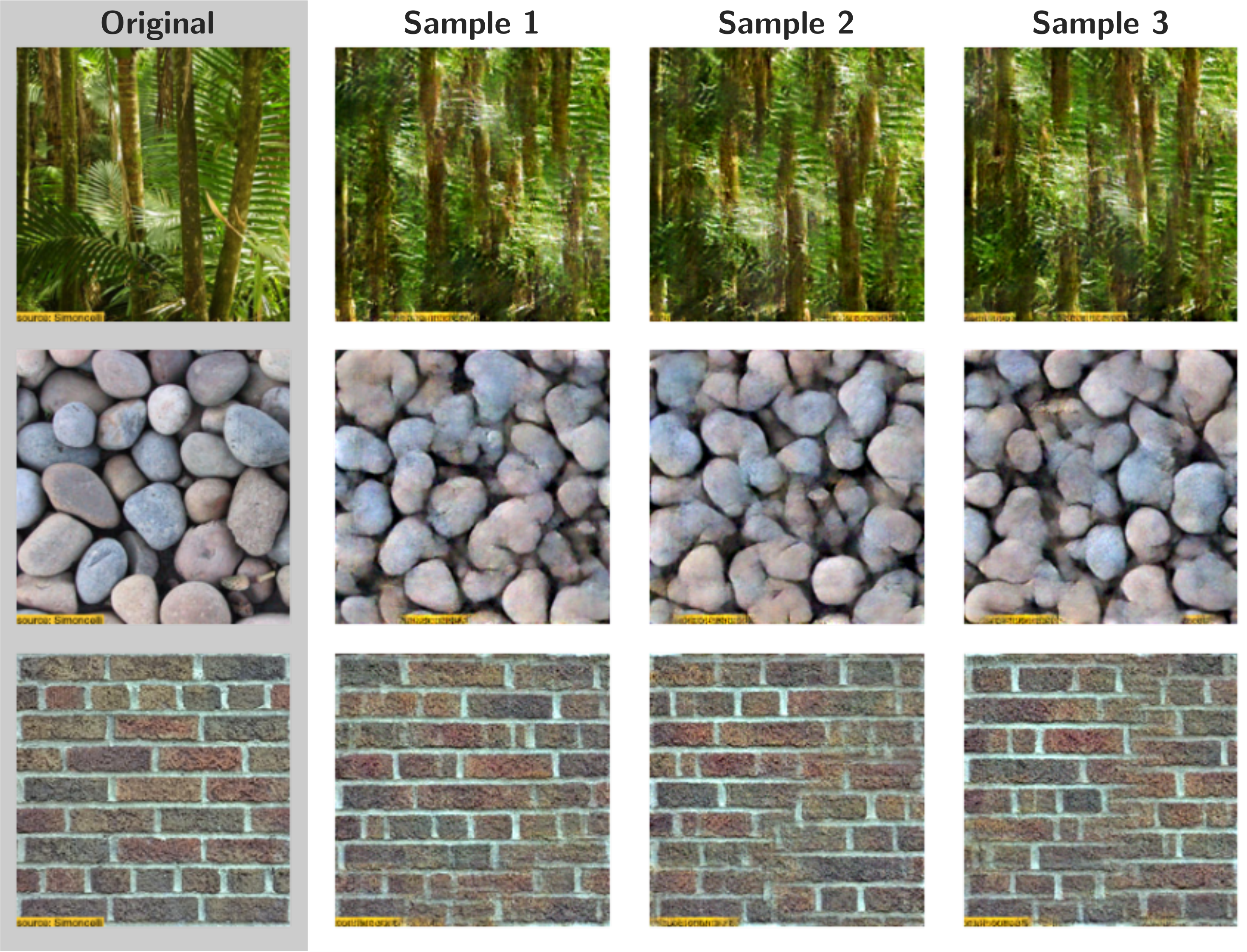}}
  \caption{Each row shows the reference texture (left, gray background) and three samples that were synthesized from different (random) initial images using our multi-scale model. Most importantly, the multi-scale model generates samples that are perceptually different. All three textures are taken from Portilla and Simoncelli, \cite{Portilla:2000}.}
  \label{fig:multi-samples}
\end{figure}
In Fig. \ref{fig:confusion} we quantify the quality of two very simple random single-layer texture models. The \textit{single-scale} model employs 1024 feature maps of size $11\times 11$ that are drawn from a zero-mean uniform distribution. The \textit{multi-scale} model employs random filters on multiple scales ranging from $3\times 3$ up to $55 \times 55$ pixels (see sec. \ref{sec:cnn}). A good texture representation should be similar for patches taken from the same texture, and very distinctive for patches from different textures. To test this, we sample 10 random $300\times 300$ patches from 10 different textures, compute the model representations (i.e. the Gram matrix) on each patch and evaluate the relative squared distance \eqref{eq:gramdist} between them. We then plot the median distance between patches of two textures as a confusion matrix, Fig. \ref{fig:confusion}. For comparison, we also plot the confusion matrix for the raw pixel values as well as for the VGG-model \cite{gatys:2015a}. The latter shows a clear distinction between within-class and between-class patches, which is completely lacking in the raw pixel space. More surprisingly, however, is the confusion matrix in the random single-layer models: their distinction of patches is on par with VGG. In other words, the texture parametrization in random shallow networks seems similarly suited to measure the perceptual difference between two patches. This intriguing finding suggests that the astonishing perceptual quality of textures synthesized from the VGG model is not, as has been thought, the result of the very specific, supervisedly trained multi-layer representation. As a consequence, images synthesized from the single-layer models should perform similarly to Gatys et al. \cite{gatys:2015a}.

In Fig. \ref{fig:filter-samples} we show textures synthesised from the random single- and multi-scale models, as well as eight other non-random single-layer models for three different source images (top left). For comparison, we also plot samples generated from the VGG model by Gatys et al. \citep{gatys:2015a} (bottom left). There are roughly two groups of models: those with a small number of feature maps (363, top row), and those with a large number of feature maps (3267, bottom row). Only the multi-scale model employs 1024 feature maps. Within each group, we can differentiate models for which the filters are unsupervisedly trained on natural images (e.g. sparse coding filters from k-means), principally devised filter banks (e.g. 2D Fourier basis) and completely random filters (see sec. \ref{sec:cnn} for all details). All single-layer networks, except for multi-scale, feature $11\times 11$ filters. Remarkably, despite the small spatial size of the filters, all models capture much of the small- and mid-scale structure of the textures, in particular if the number of feature maps is large. Notably, the scale of these structures extends far beyond the receptive fields of the single units (see e.g. the pebble texture). We further observe that a larger number of feature maps generally increases the perceptual quality of the generated textures. Surprisingly, however, completely random filters perform on par or better then filters that have been trained on the statistics of natural images. This is particularly true for the multi-scale model that clearly outperforms the single-scale models on all textures. The captured structures in the multi-scale model are generally much larger and often reach the full size of the texture (see e.g. the wall).

The perceptual quality of the textures generated from models with only a single layer and random filters is quite remarkable and surpasses previous state of the art parametric methods like Portilla and Simoncelli \cite{Portilla:2000}. The multi-scale model often rivals and sometimes even outperforms the current state of the art \cite{gatys:2015a} as we show in Fig. \ref{fig:samples} where we compare samples synthesized from 20 different textures for the random single- and multi-scale model, as well as VGG. The multi-scale model generates very competitive samples in particular for textures with extremely regular structures across the whole image (e.g. for the brick wall, the grids or the scales). In part, this effect can be attributed to the more robust optimization of the single-layer model that is less prone to local minima then the optimization in deeper models. This is exemplified in the grid structures, for which the VGG-based loss is paradoxically lower for samples from the multi-scale model then for the VGG-based model (which directly optimized the VGG-based loss). This suggests that the naive synthesis performed here favors images that are perceptually similar to the reference texture and thus looses variability (see sec. \ref{sec:discussion} for further discussion). Nonetheless, samples from the single-layer model still exhibit large perceptual differences, see Fig. \ref{fig:multi-samples}.
\afterpage{%
\begin{figure}[t]
  \centering
      \makebox[\textwidth][c]{\includegraphics[width=1\textwidth]{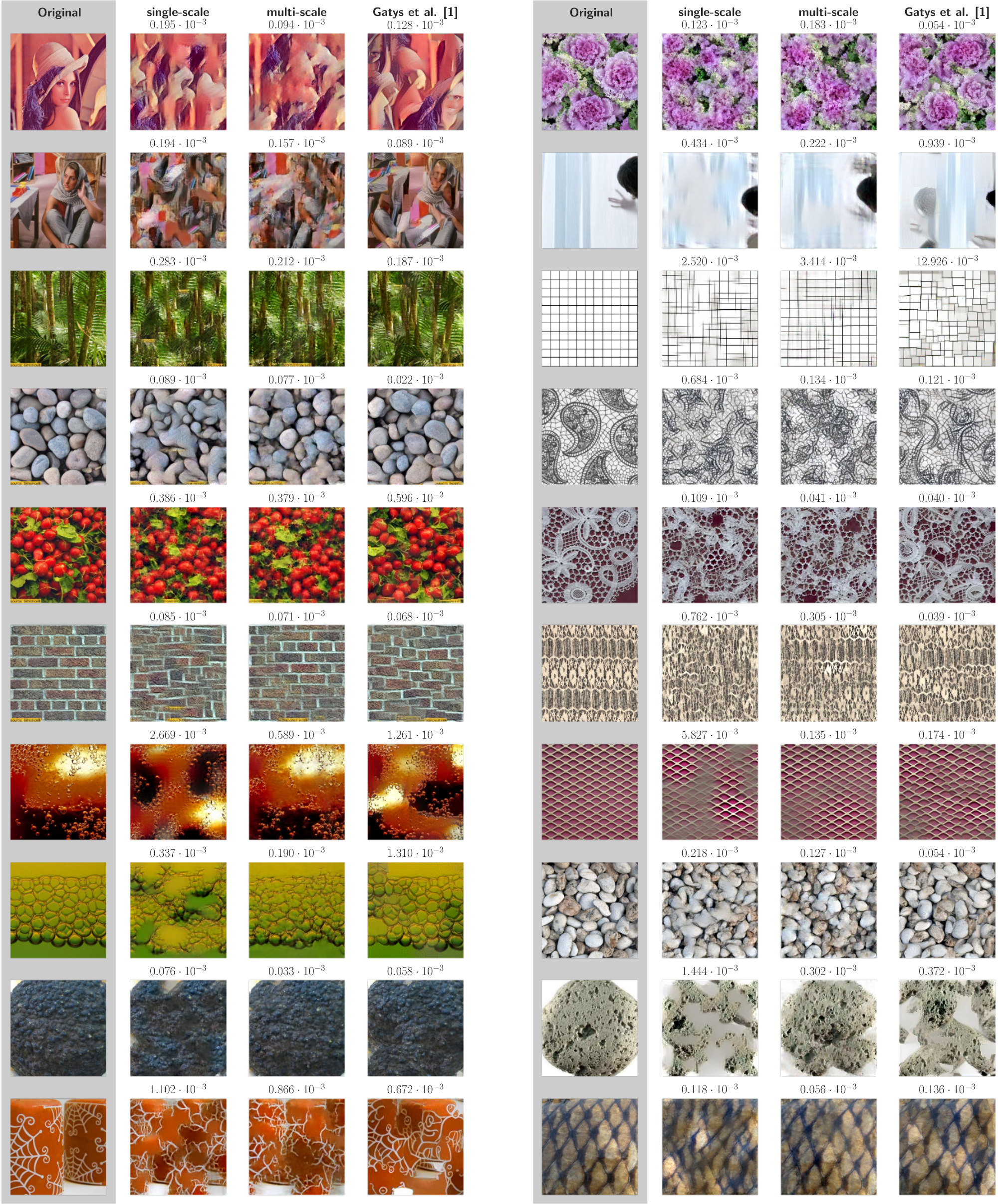}}
  \caption{Each row shows the reference texture (left, gray background) and three samples that were synthesized from different (random) initial images using three different models: single-layer network with 1024 feature maps and random 11x11 filters; multi-scale single layer network with filters of sizes $f \times f$, where $f = \{3, 5, 7, 11, 15, 23, 37, 55\}$ and 128 feature maps correspond to filters of each size; and the VGG-based model \cite{gatys:2015a}. Numbers above figures show the values of the normalized VGG-loss \eqref{eq:VGGloss} for corresponding textures.}
  \label{fig:samples}
\end{figure}
\clearpage
}
The VGG-based loss \eqref{eq:VGGloss} appears to generally be an acceptable approximation of the perceptual differences between the reference and the synthesized texture. Only for a few textures, especially those with very regular men-made structures (e.g. the wall or the grids), the VGG-based loss fails to capture the perceptual advantage of the multi-scale synthesis.

\section{Discussion}
\label{sec:discussion}

In this paper we demonstrated a new parametric texture model based on a single-layer convolutional network with random filters. We showed that the model is able to qualitatively capture the perceptual differences between natural textures. Samples from the model often rival and sometimes even outperform the current state of the art \citep{gatys:2015a} (Fig. \ref{fig:samples}, third vs fourth row), even though the latter relies on a high-performance deep neural network with features that are tuned to the statistics of natural images. This finding suggests that neither the hierarchical texture representation, nor the trained filters are critical for high-quality texture synthesis. Instead, both aspects rather seem to serve as fine-tuning of the texture representation.

Our results clearly demonstrate that Gram matrices computed from the feature maps of convolutional neural networks generically lead to useful summary statistics for texture synthesis. The Gram matrix on the feature maps transforms the representations from the convolutional neural network into a stationary feature space that captures the pairwise correlations between different features. If the number of feature maps is large, then the local structures in the image are well preserved in the projected space and the overlaps of the convolutional filtering add additional constraints. At the same time, averaging out the spatial dimensions yields sufficient flexibility to generate entirely new textures that differ from the reference on a patch by patch level, but still share much of the small- and long-range statistics.

The success of shallow convolutional networks with random filters in reproducing the structure of the reference texture is remarkable and indicates that they can be useful for parametric texture synthesis. Besides reproducing the stationary correlation structure of the reference image ("perceptual similarity") another desideratum of a texture synthesis is to exhibit a large variety between different samples generated from the same given image ("variability"). Hence, synthesis algorithms need to balance perceptual similarity and variability. This balance is determined by a complex interplay between the choice of summary statistics and the optimization algorithm used. For example the stopping criterion of the optimization algorithm can be adjusted to trade perceptual similarity for larger variability. 

In this paper we focused on maximizing perceptual similarity only, and it is worth pointing out that additional efforts will be necessary to find an optimal trade-off between perceptual similarity and variability. For the synthesis of textures from the random models considered here, the trade-off leans more towards perceptual similarity in comparison to Gatys et al. \cite{gatys:2015a}(due to the simpler optimization) which also explains the superior performance on some samples. In fact, we found some anecdotal evidence (not shown) in deeper multi-layer random CNNs where the reference texture was exactly reconstructed during the synthesis. From a theoretical point of view this is likely a finite size effect which does not necessarily constitute a failure of the chosen summary statistics: for finite size images it is well possible that only the reference image can exactly reproduce all the summary statistics. Therefore, in practice, the Gram matrices are not treated as hard constraints but as soft constraints only. More generally, we do not expect a perceptual distance metric to assign exactly zero to a random pair of patches from the same texture. Instead, we expect it to assign small values for pairs from the same texture, and large values for patches from different textures (Fig. \ref{fig:confusion}). Therefore, the selection of constraints is not sufficient to characterize a texture synthesis model but only determines the exact minima of the objective function (which are sought for by the synthesis). If we additionally consider images with small but non-zero distance to the reference statistics, then the set of equivalent textures increases substantially, and the precise composition of this set becomes critically dependent on the perceptual distance metric.

Mathematically, parametric texture synthesis models are described as ergodic random fields that have maximum entropy subject to certain constraints \cite{zhu:97,bruna:2013,zhu:00} (MaxEnt framework). Practical texture synthesis algorithms, however, always deal with finite size images. As discussed above, two finite-size patches from the same ergodic random field will almost never feature the exact same summary statistics. This additional uncertainty in estimating the constraints on finite length processes is not thoroughly accounted for by the MaxEnt framework (see discussion on its ``ad hockeries'' by Jaynes \cite{jaynes:1982}). Thus, a critical difference of practical implementations of texture synthesis algorithms from the conceptual MaxEnt texture modeling framework is that they genuinely allow a small mismatch in the constraints. Accordingly, specifying the summary statistics is not sufficient but a comprehensive definition of a texture synthesis model should specify:
\begin{enumerate}
	\item A metric $d({\bf x}, {\bf y})$ that determines the distance between any two arbitrary textures ${\bf x}, {\bf y}$.
	\item A bipartition $P_{\bf x}$ of the image space that determines which images are considered perceptually equivalent to a reference texture ${\bf x}$. A simple example for such a partition is the $\epsilon$-environment $U_\epsilon({\bf y}) := \{{\bf y} : d({\bf y}, {\bf x}) < \epsilon\}$ and its complement.
\end{enumerate}
This definition is relevant for both under- as well as over-constrained models, but its importance becomes particularly obvious for the latter. According to the Minimax entropy principle for texture modeling suggested by Zhu et al \cite{zhu:97}, as many constraints as possible should be used to reduce the (Kullback-Leibler) divergence between the true texture model and its estimate. However, for finite spatial size, the synthetic samples become exactly equivalent to the reference texture (up to shifts) in the limit of sufficiently many independent constraints. In contrast, if we explicitly allow for a small mismatch between the summary statistics of the reference image and the synthesized textures, then the set of possible textures does not constitute a low-dimensional manifold but rather a small volume within the pixel space.

Taken together we have shown that simple single-layer CNNs with random filters can serve as the basis for excellent texture synthesis models that outperform previous hand-crafted synthesis models and may even rival the current state of the art. This finding repeals previous observations that suggested a critical role for the multi-layer representations in trained deep networks for natural texture generation. On the other hand, it is not enough to just use sufficiently many constraints as one would predict from the MaxEnt framework. Instead, for the design of good texture synthesis algorithms it will be crucial to find distance measures for which the $\epsilon$-environment around the reference texture leads to perceptually satisfying results. In this way, building better texture synthesis models is inherently related to better quantitative models of human perception. 

\vfill
\pagebreak

\bibliographystyle{ieeetr}
\bibliography{biblio}

\end{document}